\pdfoutput=1

\documentclass[11pt]{article}

\usepackage[final]{acl}

\usepackage{times}
\usepackage{latexsym}

\usepackage[T1]{fontenc}

\usepackage[utf8]{inputenc}

\usepackage{microtype}

\usepackage{inconsolata}

\usepackage{graphicx}

\usepackage{multirow}
\usepackage{xltabular}
\usepackage{diagbox}
\usepackage{makecell} 
\usepackage{tabularx}

\newcommand{\role}[1]{\textit{#1}}
\newcommand{\argZero}{\role{Arg0}}
\newcommand{\argOne}{\role{Arg1}}
\newcommand{\argtwo}{\role{Arg2}}
\newcommand{\ChainOfThought}{chain-of-thought}
\newcommand{\location}{\role{Location}}
\newcommand{\instrument}{\role{Instrument}}
\newcommand{\simplePrompting}{Simple Prompting}
\newcommand{\stepByStepPrompting}{Step-by-Step Prompting}
\newcommand{\experiment}[1]{Exp.{#1}}

\newcommand{\gptTurbo}{GPT4-Turbo}
\newcommand{\gptFourOne}{GPT4.1}

\newcommand{\llama}{Llama}
\newcommand{\qwen}{Qwen}

\newcommand{\rhoSymbol}{\textbf{$\rho$}}
\newcommand{\llm}{LLM}
\newcommand{\llms}{LLMs}
\newcommand{\FerIns}{Fer-Ins}
\newcommand{\FerLoc}{Fer-Loc}
\newcommand{\prompt}[1]{\texttt{\small #1}}
\usepackage{enumitem}

\definecolor{darkgreen}{RGB}{0,150,0}

\usepackage{todonotes}
\newcommand{\ignore}[1]{}

\newcommand{\anon}[1]{(anonymized)}

\title{Uncovering Autoregressive LLM Knowledge of Thematic Fit \\ in Event Representation}

\author{ Safeyah Khaled Alshemali \\
  Imperial College London \\
  London, UK  \\
  \texttt{ska24@ic.ac.uk} 
  \And
  Daniel Bauer \\
  Columbia University \\
  New York, NY, USA \\
  \texttt{db2711@columbia.edu} 
  \And
  Yuval Marton \\
  University of Washington \\ WA, USA \\
  \texttt{ymarton@uw.edu}}

\begin{document}
\maketitle
\begin{abstract}
The thematic fit estimation task measures semantic arguments' compatibility with a given semantic role for a given predicate.
We investigate if autoregressive LLMs have consistent, expressible knowledge of event arguments' thematic fit by experimenting with various prompt designs, manipulating input context, reasoning, and output forms.
We set a new state-of-the-art on thematic fit benchmarks, but show that closed and open weight LLMs respond differently to our prompting strategies: Closed models achieve better scores overall and benefit from multi-step reasoning, but they perform worse at filtering out generated sentences incompatible with the given predicate, role, and argument.
Our analysis shows that lemma tuple input and sentence input result in surprisingly different thematic fit score distributions.

%
%

\end{abstract}

\section{Introduction}\label{sec:intro}
Do large language models (LLMs) have consistent, expressible knowledge 
of event arguments' thematic fit to a semantic role for a given predicate? 

\paragraph{Thematic fit}
is the level of  compatibility between the predicate (typically a verb), its argument (typically a noun phrase), and a  semantic role assigned to the argument.
For example, the verb ‘eat’ invokes a set of potential arguments for specific semantic roles. If the role is \role{Agent} (or \argZero\  in PropBank; see Section \S\ref{sec:Datasets}), then ‘people’, ‘cat’,  etc., would be a good fit, as opposed to ‘pizza’, ‘apple’, etc., as the latter group would better fit a \role{Patient} role  (often denoted as \argOne\  in PropBank). If the role is \location, then ‘restaurant’, ‘kitchen’,  etc., would be a good fit due to their properties. If the role is \instrument, then ‘fork’, ‘knife’, etc. would be a good fit due to their use as tools while eating.
Psycholinguistic experiments show that humans rely on such generalized, prototypical knowledge about events and their typical participants during language understanding, early in the process and prior to syntactic interpretation \citep{mcrae2009people,bicknell2010effects}.

\paragraph{Thematic fit estimation (TFE)} applies computational methods and event representations derived from data to mimic 
this type of human knowledge.
This is challenging due to the lack of direct labeled training data. Thematic fit norms (see §\ref{sec:Datasets}) summarize human ratings and provide a benchmark for evaluation, but are not large enough for direct supervised training. Instead, TFE models are traditionally trained on SRL datasets as proxy data.\footnote{TFE is related to
semantic role labeling (SRL). 
SRL  outputs the semantic frames in the input sentence, using frameworks such as PropBank \citep{kingsbury2002treebank} or FrameNet \citep{johnson2016framenet}. The output includes the predicates, arguments, and their semantic roles in the frame. TFE takes the predicates, arguments, and roles as input, and outputs a score. Understanding the thematic fit of individual arguments/roles is thought to be required for SRL, but is not usually modeled explicitly in SRL systems.  For theoretical and historical details, see 
\citet{gruber1965studies,fillmore1968lexical,dowty1991thematic,parsons:1990,mcrae1998modeling}  \textit{inter alia}.} 
Recently, pretrained language models have shown state-of-the-art performance without task-specific supervised training on many computational semantics tasks, such as synonym judgment \citep{levy2017semantic}, similarity judgments \citep{hill2015simlex}, and more. 
However, on TFE, these models, including masked LMs such as BERT  \citep{devlin2019bert}, have so far only yielded modest improvements over previous approaches \citep{pedinotti2021cat}. 
In this work, we examine if autoregressive LLMs, such as GPT \citep{classicalGPT, achiam2023gpt}, can produce better results via prompting. We design our experiments to elicit linguistic knowledge along the following three axes (see \S\ref{sec:ExperimentDesign} for details): 

\paragraph{Axis 1: Reasoning \small{(Simple vs. \stepByStepPrompting)}} \label{subsec:axis1}
\llms\ perform remarkably well in challenging NLP tasks such as Planning \citep{wang2022towards}, automatic scoring \citep{lee2024applying}, and question answering \citep{wang2023keqing}, when using  ‘\ChainOfThought\  prompting’  \citep{wei2022chain}. This technique breaks a problem into sub-steps toward answering the core question. It seems to help \llms\ in a similar way a scratchpad would help someone solve a math exercise.\footnote{It proved so influential that several leading labs have developed `thinking' models with `test-time compute', incorporating this approach in the post-training and settings.}
We view TFE as a linguistic \textit{reasoning} task because it can be broken into clear sub-steps.
Thus, we compare the \llms' performance using a “naive”  single-prompt approach (simply asking the model to directly score a given lemma tuple; \textbf{\simplePrompting}\ hereafter),   with the \ChainOfThought\  approach (\textbf{\stepByStepPrompting}\   hereafter).

\paragraph{Axis 2: Input \small{(Generated Sentences vs. Lemma Tuples)}} \label{subsec:axis2}
The TFE norm data comes in a lemma tuple format containing a predicate lemma, lemma of the syntactic head of the predicate's argument, its thematic role, and the argument's thematic fit score (an average of several human raters' scores). For example, $\langle$predicate: eat, argument: pizza, role: \argZero/\role{Agent}, score:1.3$\rangle$. Since \llms\ were pretrained on text rather than such tuples, we hypothesize that they could benefit from seeing full sentences that use the predicate and argument in the given role. We prompt the LLM to generate such sentences and use a novel approach for filtering out semantically incoherent and incompatible sentences.\footnote{Upon a close reading, we find that the human raters were indeed provided full sentences for the rating \citep{mcrae1998modeling}, but this fact seems to have been lost over the years, and only the lemma tuple format has been used (\citealp{tilk2016event}; \citealp{hong2018learning}; \citealp{muthupari2022s}). We close here a full circle, bringing the conditions closer to the original.\label{footnote3}} 

\paragraph{Axis 3: Output \small{(Predefined Categories vs. Numeric Score)}} \label{subsec:axis3}
The thematic fit ratings in the TFE norms are Likert scale ratings (following \citet{mcrae1998modeling}) averaged over 7 human raters. A straightforward evaluation would prompt the model to output numeric scores as well. 
However, \llms\  were designed to handle textual data, so outputting numeric scores may lead to inconsistent results.  Therefore, we also prompt the \llms\ to rate the thematic fit according to predefined categories (“Low”, “High”, etc.), which we then convert to corresponding predefined numeric values for evaluation purposes (see Section~\S\ref{sec:Eval}).\footnote{We leave for future work whether it is useful to use \llms' logprobs for this task, and how to best do that.}

\paragraph{}
Our experiments compare the performance of two closed-weight LLMs (\gptFourOne\ and \gptTurbo) and two open LLMs (Llama3.2 and Qwen2.5). We show  closed LLMs achieve state-of-the-art results and demonstrate the benefits of \stepByStepPrompting\ on the TFE task. In contrast, open LLMs demonstrate worse zero-shot performance on this task, but may have an edge on  filtering out  generated sentences if incompatible with the specified  predicate, role, and argument. 

\paragraph{Contributions}
Our main contributions are:
\begin{itemize}[itemsep=2pt, topsep=2pt, parsep=3pt, partopsep=1pt, leftmargin=*]
  \setlength\itemsep{-.2em}
  \item First study to prompt autoregressive \llms\ in order to  estimate thematic fit from output tokens.
  \item New state-of-the-art thematic fit estimation.
  \item Comparison of prompting techniques along three axes: reasoning form, input form (with/out LLM-filtered generated sentences), and output form.
  \item An analysis of related strengths and weaknesses of various open and closed weight LLMs.
\end{itemize}

\noindent Source code is available at: {\url{https://github.com/SafeyahShemali/LLM_Thematic_Fit_25}} 

\section{Related Work} \label{sec:RelatedWork}
TFE is closely related to Selectional Association \cite{Resnik1996SelectionalCA},
which is an information-theoretic measure of the difference in the probability that some noun class will appear as a specific syntactic argument of the given predicate, compared to the general probability of this class.\footnote{It requires a taxonomy (WordNet) for the noun classes and a syntactically parsed corpus for estimating probabilities. 
Thematic fit focuses on the cognitive load of understanding an utterance with the given predicate and argument(s) in specific thematic roles, 
\textit{post-hoc}.
TFE also uses a corpus, but with the advent of LLMs, no explicit parses or noun classes are needed, hence it is more applicable and fine-grained.
}

Early distributional approaches to the TFE task compute the similarity between vector representations of an argument (usually for the argument's syntactic head word, obtained from a distributional semantic model or static word embeddings) and a prototype vector for the predicate/role pair \citep{Pad2007FlexibleCM,erk2010flexible,baroni2010distributional,santus2017,sayeed2015exploration, greenberg2015improving}. Our first baseline (hereafter {\bf BG}) is the distributional approach by {\citeauthor{greenberg2015improving}, which is based on the syntactically structured distributional semantic model by \citeauthor*{baroni2010distributional}.
Following approaches use neural networks to learn event representations. \citet{tilk2016event} describe a {\it non-incremental role filler} model (NN RF, baseline {\bf B0}) trained to predict role fillers, given as input a predicate, target role, and optionally other contextual roles and their fillers. \citet{hong2018learning} improve over \citet{tilk2016event} by adding role prediction as a multi-task objective, and improving pooling of context roles/fillers (ResRoFA-MT, baseline {\bf B1}). \citet{Marton2021ThematicFB} replicate \citet{hong2018learning}'s model and train it on higher quality silver labels, experimenting with the quality/quantity tradeoff of annotations (baseline {\bf B2}). \citet{muthupari2022s} focus on evaluating the capability of various models, or model elements, to capture thematic fit. They compare initializing the \citet{hong2018learning} model with random vectors vs. pre-trained GloVe embeddings \citep{pennington2014glove},  and other modifications. We add three of their higher performing models to our baselines: ({\bf B3a}) Glove-shared-not-tuned, ({\bf B3b}) Glove-shared-tuned, ({\bf B3c}) Glove-shared-tuned-20\%M. B3a and B3b report the average scores over several runs, whereas B3c is identical to B3b except for reporting the maximum score over several runs, all trained on the same 20\% subset of the data. 

Contemporary work using pretrained LLMs (without any additional supervised training) for TFE either examines if these models can differentiate plausible from implausible sentences (``the teacher bought the laptop'' vs. ``the laptop bought the teacher''; \citet{kauf2023event}), or used BERT-style models to predict role fillers for a masked argument in the context of a generated sentence \cite{metheniti2020relevant,pedinotti2021cat}. Notably, \citeauthor{pedinotti2021cat} only report a moderate improvement over a distributional approach that uses a high quality structured distributional model \citep{chersoni2019structured}.  
\citet{vassallo2018event} introduce DTFit, a dynamic evaluation of thematic fit as more words are added to a sentence already containing an \role{Agent} and a predicate (verb). Like us, they use autoregressive \llms\, but focus on \role{Agent} with other roles, use \llm\ logprobs, not using tuples as input, and are constrained to specific sentence structures.
\citet{testa2023we} construct ELLie, a dataset including various types of elliptical constructions with typical, atypical, anomalous completions, based on \citet{culicover2005simpler} along with DTFit, to examine the interaction between thematic fit and the ability of LLMs to resolve verbal ellipsis.
\citet{kauf2024log} compare base and instruction-tuned LLMs' ability of estimating event plausibility.

We are the first to compare autoregressive \llms' TFE over $\langle$predicate, argument, role$\rangle$ tuples vs. generated (and filtered) sentences; simple prompting vs. chain-of-thought reasoning; and prompting for numeric output vs. pre-defined categories.

\begin{figure*}[t]
    \centering
    \resizebox{.8\linewidth}{!}{
    \includegraphics[width=\textwidth]{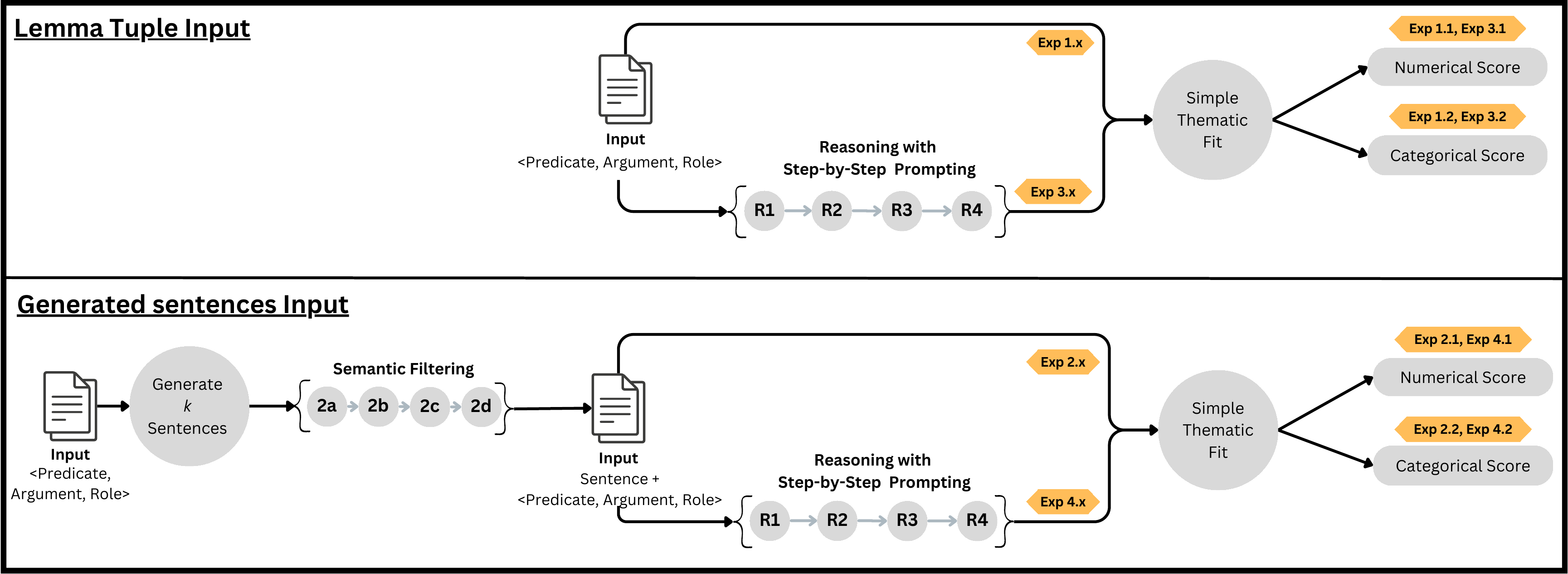}
    }
    \caption{Experiment Method. For more details of the \stepByStepPrompting\ and Semantic Filtering, see \S\ref{sec:reasoning}. The output of \experiment{3.x} - \experiment{4.x} contains the score's justification in addition to the score.}
    \label{fig:overviewExp}
\end{figure*}

\begin{table}[b]
    \centering
    \small
    \resizebox{\columnwidth}{!}{
     \begin{tabular}[3]{|c|c|l|l|} 
     \hline
     \textbf{Reasoning Form} & \textbf{Input Form} & \textbf{Output Form} & \textbf{Exp}\\ 
     \hline 
     \multirow[m]{4}{*}{\simplePrompting} &\multirow[m]{2}{*}{Head Lemma Tuples} & Numerical & 1.1\\\ 
     && Categorical & 1.2\\ 
     \cline{2-4} 
     &\multirow[m]{2}{*}{Generated Sentences} & Numerical & 2.1\\ 
     && Categorical & 2.2\\
     \hline
     \multirow[m]{4}{*}{\stepByStepPrompting} & \multirow[m]{2}{*}{Head Lemma Tuples} &  Numerical & 3.1\\ 
     &&  Categorical & 3.2\\ 
     \cline{2-4} 
     & \multirow[m]{2}{*}{Generated Sentences} &  Numerical & 4.1\\ 
     &&  Categorical & 4.2\\ 
     \hline
    \end{tabular} }
    \caption{Overview of the experiments. }
    \label{tab:OverviewT}
\end{table}

\section{Experiments} \label{sec:Experiments}
The experiments were structured along three axes (see \S\ref{sec:intro}) with two possible values each. Therefore, we conduct $2^3=8$ experiments. We refer to them as `Exp $n.m$' (see Table~\ref{tab:OverviewT}). Figure~\ref{fig:overviewExp} shows a schematic overview of the different methods.
Each experiment has a prompting template according to the reasoning, input, and output settings, as detailed in Appendix \S\ref{sec:ApdxPromptDesign}. 

\subsection{Method: Reasoning, Input, and Output} \label{sec:ExperimentDesign}
\paragraph{Axis 1: Reasoning  \small{(\simplePrompting\  vs. \stepByStepPrompting)}}\label{sec:reasoning}
The first axis is concerned with the effectiveness of  \stepByStepPrompting\  (\experiment{3.x and 4.x}) compared to \simplePrompting\  (\experiment{1.x and 2.x}) in evaluating the thematic fit. In the \simplePrompting\  approach we use a single zero-shot prompt to provide the model with a predicate, argument, and semantic role, and ask it to rate the thematic fit. 
 \stepByStepPrompting\    decomposes the task into multiple steps, each inferring properties needed to be considered for scoring the thematic fit of the given tuple $\langle$predicate, argument, semantic role$\rangle$. 
 These steps are:
\textbf{R1.} listing the argument’s (salient or relevant) properties, \textbf{R2.} listing the (prototypical or necessary) properties required for 
the argument in the role “assigned” by the given predicate, \textbf{R3.} listing the (suitable or likely) roles for the argument given predicate,  \textbf{R4.} listing the argument's properties that fit well or are missing for the specified role given the predicate,
\textbf{R5.} and finally, estimating the argument’s thematic fit to the assigned role, given these four lists. 
 The last step is similar to the `simple thematic fit prompt’ in the \simplePrompting\  setting. 

\paragraph{Axis 2: Input \small{(Lemma Tuples vs. Generated Sentences)}}\label{sec:input}
The input in each experiment can take one of two forms. The basic form (available in the TFE norm data and used in previous work including our baselines) is a $\langle$predicate, argument, semantic role$\rangle$ tuple, where the predicate is a verb lemma and the argument is represented by the lemma form of its syntactic head. Adjuncts, subordinate clauses, etc., are ignored. The semantic role is taken from either PropBank or VerbNet. For example, for the sentence “I ate a pizza with my friends” we would see $\langle$eat, I, \role{Agent}$\rangle$ or $\langle$eat, pizza, \role{Arg1}$\rangle$. 
The \textbf{Lemma Tuples} setting (\experiment{1.x and 3.x}) uses only the lemmas, as in previous work\footnote{In  \citet{mcrae1998modeling}, the  raters were given full sentences, but they were not made public as far as we know (see footnote~\ref{footnote3}).}
In the \textbf{Generated Sentences} setting (\experiment{2.x and 4.x}), a sentence generation step happens before the other prompts. 
We hypothesize that providing more context,  similar  to what the LLM presumably was trained on, will improve performance. Sentence generation consists of two steps: 
\textbf{1.~Generation:} prompting the \llm\ to generate $k=5$ candidate sentences with the given predicate, argument, and role (see \S\ref{sec:gensentenceApx}), and 
\textbf{2.~Filtering:} instructing the \llm\ to confirm each generated sentence's semantic cohesion and its use of the specified predicate and argument in its specified semantic role.  
The second step is needed, as the models (as tested in preliminary trials) seem to often generate semantically incoherent sentences, or use the given argument not in the given semantic role. Each sentence is given three trials to pass the filtering process, which consists of 4 waterfall Yes/No steps: \textbf{(2a)} is the given sentence semantically coherent, \textbf{(2b)} does the sentence contain the predicate, \textbf{(2c)} does the sentence contain the argument as the given predicate's argument, and \textbf{(2d)} is the argument in the required role for the predicate (see \S\ref{sec:gensentenceApx}).
The sentence must pass all steps of the filtering process in at least one trial, or it will be filtered out.  The final fit score is an average over the remaining sentences' scores. If steps (1) or (2) fail, we back off to using the lemma tuples.

\paragraph{Axis 3: Output \small{(Categorical vs. Numeric  Scoring)}} \label{sec:output-form}\label{sec:output}
Each experiment may generate one of two forms of output: a numeric thematic fit score (\experiment{x.1} in Table~\ref{tab:OverviewT}) or a predefined thematic fit category (\experiment{x.2}). The numeric thematic fit score (numeric score, hereafter) is a numeric value in the range [0,1]. The predefined thematic fit categories (categorical score, hereafter) are {`Unlikely/Impossible', `Low', `Medium', `High', or `Perfect'}. These output score categories are then converted to the following numeric values: (0.0, 0.25, 0.5, 0.75, 1.0), respectively, in a post-processing script, for evaluation purposes (see \S\ref{sec:Eval}). In \experiment{3.x} and \experiment{4.x}, the model is asked to generate a justification for the score it assigns to the input, based on its previous \stepByStepPrompting\ reasoning output.


\subsection{Datasets and Evaluation} \label{sec:Datasets}\label{sec:Eval}

\paragraph{Training set:} We used pre-trained LLMs (see Section \S\ref{Models}) “off the shelf” without modification. 
In an ideal world, we would be able to point the reader to the full pre-training data. We did not add psycholinguistic (or other) labeled data to the training set, nor did we fine-tune the LLMs.

\paragraph{Test sets:} We used the following datasets to evaluate our methods.\footnote{These datasets are too small to be split to training, validation and test sets, as is common in machine learning. We follow here previous TFE literature.}
Each of these test sets comprises a list of $\langle$predicate, argument, semantic role, score$\rangle$ tuples. For example, $\langle$eat, restaurant, \location, 0.9$\rangle$. The first two are the representative lemma form of the syntactic head of the predicate and argument (in some sentence), the semantic role is as listed below, and the score is the ground truth (human rating). These tuples are given to the LLMs as input, except for the score, which is used to evaluate the models' output (details below).

{\citet{mcrae1998modeling}}, hereafter \textbf{`McRae',} contains $1,444$ predicate-argument-role-rating tuples,  developed with the participation of 120 native English speakers undergraduate students at the University of Rochester. Students were asked to rate the role typicality (thematic fit) of a predicate (verb) and the argument (noun) in sentence fragments on a Likert scale of 1 to 7.
The public dataset does not contain the sentence fragments. The roles are PropBank’s \argZero\  and \argOne, which are roughly equivalent to \role{Agent} and \role{Patient}, respectively \citep{kingsbury2002treebank}. We removed eight duplicates. 

{\citet{pado2006modelling}}, hereafter \textbf{`Pado'}, contains 414 tuples in the same form as \citet{mcrae1998modeling}; however, the role could also be \argtwo. 
To prevent bias in rating the same verb-noun with different roles, one hundred native English speaker volunteers judged tuples in four randomly ordered lists.

{\citet{ferretti2001integrating}} evaluate thematic fit for the roles \instrument\ (248 tuples; hereafter, the \textbf{\FerIns}\ dataset) and \location\ (277 tuples; hereafter, \textbf{\FerLoc} dataset). Scores were collected from fifty-eight students. The rating scale is the same as in \citet{mcrae1998modeling}. For \instrument, the participants were asked, “How common is it for someone to use each of the following to perform the \textit{eating} action?” For  \location, the question was, “How common is it for someone to \textit{eat} in each of the following locations?”. We removed  indefinite articles in \FerLoc\ arguments,  in order to consider only the lemma of the syntactic head of the argument (not the phrase).

\paragraph{Evaluation:}\label{SpearmanEval}  
For each test set, we used 
Rank Correlation Coefficient (\rhoSymbol) \citep{spearman:1904} between two lists:  
the test set, sorted by the average human ratings for each item, and the corresponding model output. Categorical output was converted to numeric values, so it can be sorted similarly too before conducting this evaluation (see \S\ref{sec:output-form}). 

\subsection{Hyperparameters}\label{sec:hyperparams}

We conducted a preliminary study to determine good temperature and top\_p parameter values for the \llms. We used GPT-3.5   on a subset of the \FerIns\ dataset (20 samples $\sim$ 8\%) and ran a parameter sweep over temperature values (0.0, 0.5, 0.9) and top\_p values (0.5, 0.7, 0.95). Temperature values were chosen near-uniformly across the scale to examine a wide range of the model creativity setting. Higher values correspond to higher randomness (aka "creativity") in the output. High values of top\_p limit the randomness in the output to more probable tokens. Since we were more interested in accurate output, and less interested in creative writing for our purposes here, we tested top-p threshold values on the higher half of the scale. Temperature~0.0 and top\_p~0.95 showed the highest correlation coefficient with human judgments on our sample. We set the max\_tokens parameter based on the length of the prompt and its expected output. For experiments with Lemma Tuples input, we set  max\_tokens to 100, whereas for  Generated Sentences input, we set it to 300 (for Simple prompting). As  \stepByStepPrompting\  is expected to be lengthy to clarify the underlying reasons, we set  max\_tokens to 600 there. We used these settings in all experiments.

\subsection{LLMs} \label{Models}

We used closed and open \llms, the latest available at the time of running the experiments. \textbf{Closed models} (often best performing, but less reproducible):  OpenAI gpt-4-0125-preview \citep{achiam2023gpt,openai2023gpt4turbo} with unconfirmed 1.76 trillion parameters   (hereafter \textbf{‘\gptTurbo’})  and \textit{gpt-4.1-2025-04-14} \citep{openai2025gpt41docs} with unconfirmed 1.8 trillion parameters (hereafter \textbf{`\gptFourOne'}).
\textbf{Open models} (more controlled and reproducible, even if not fully open): \textit{llama3.2:3b-instruct-q4\_K\_M} \citep{Dubey2024Llama3,meta2024llama}, a quantized version of Meta Llama3.2, 3 billion parameters (hereafter \textbf{‘\llama’})  and \textit{qwen2.5:7b-instruct-q4\_K\_M} \citep{Yang2024Qwen2.5} with 7.61 billion parameters (hereafter \textbf{‘\qwen’}). 
Closed model inference utilized the OpenAI API. Open model inference utilized Ollama \citep{Ollama2023}.

\section{Results}\label{sec:results}

\subsection{Closed Models} \label{sec:Closed Models}
Overall, our results (Table~\ref{tab:gpt4turbo}) set a new state-of-the-art for all four tasks, especially with \gptTurbo\ using \stepByStepPrompting.
Both \gptTurbo\ and \gptFourOne\ outperformed the baselines by a large margin, except for Pado (where \gptTurbo\ tied with the highest baseline, and \gptFourOne\ was weaker than most baselines).  {\gptTurbo}\ demonstrated superior performance across the board, outperforming also the newer \gptFourOne.
Along the three axes, the model performance differed as follows:

\paragraph{Axis 1: Reasoning Form}
\simplePrompting\ with \gptTurbo\  outdid all baselines (except Pado), while results with \gptFourOne\ were mixed, in spite of this model being newer. 

For both \gptTurbo\ and \gptFourOne, \stepByStepPrompting\ (Exp.3.x and 4.x) yielded the best scores, demonstrating the effectiveness of this approach. For \gptTurbo, absolute improvement over the strongest baseline was  $.23$ for \FerLoc, $.22$ for \FerIns, $.22$ for McRae, and a tie with Pado.

\paragraph{Axis 2: Input Form}
For both closed models, using Generated Sentences (Exp.2.x and 4.x) rather than Lemma Tuples was detrimental. 
The only exception was for Pado with \gptTurbo, where  Generated Sentences with \stepByStepPrompting\ resulted in our highest score for Pado.

\paragraph{Axis 3: Output Form}
Using predefined categories compared to numeric output (\experiment{x.2} vs. \experiment{x.1}) had zero, or small mixed effect, contrary to our hypothesis.
With \gptTurbo, using predefined categorical output compared to numeric output (Exp.x.2 vs. Exp.x.1) leads to a mixed effect across the board. Regardless of input and reasoning forms, the categorical score is lower than the numerical score in \FerLoc. For \FerIns\ and Pado, the categorical score is either the same or higher than the numerical score, as the gain in \FerIns\ was .04 in Exp2 and in Pado was .04 in Exp3. Using categorical output in McRae lowers the score in general, except for Exp4, where the gain was .02.
}

\begin{table}[t!]
     \resizebox{\columnwidth}{!}{
        \begin{tabular}{|p{3cm}|c|c|c|c|}
        \hline
        \diagbox{\textbf{Model}}{\textbf{Dataset}} & \textbf{\FerLoc} & \textbf{\FerIns}  & \textbf{Pado} & \textbf{McRae} \\ \hline
        \textbf{BG:}  GSD2015 & .29 & .42 & .53 & .36 \\ \hline
        \textbf{B0:} NN RF & .44 & .45 & .52 & .38 \\ \hline
        \textbf{B1:} ResRoFA-MT & .46 & .48 & .53 & .43 \\ \hline
        \textbf{B2:}  20\% subset v2 & - & - & .43 & .44 \\ \hline
        \textbf{B3a:}  Glove-shared-not-tuned & .30 & .34 & .49 & .31 \\ \hline
        \textbf{B3b:} Glove-shared-tuned & .29 & .29 & .53 & .33 \\ \hline
        \textbf{B3c:} Glove-shared-tuned with 20\%M & .35 & .43 &\textbf{.59}& .43 \\ 
        \hline
             
        \hline
        \textbf{\experiment{1.1}} & .63          & .64          & .35           & .54 \\ \hline
        \textbf{\experiment{1.2}} & .61          & .65          & .40           & .58 \\ \hline
        \textbf{\experiment{2.1}} & .48          & .49          & .42           & .55 \\ \hline
        \textbf{\experiment{2.2}} & .46          & .54          & .45           & .56 \\ \hline
        \textbf{\experiment{3.1}} & \textbf{.69} & \textbf{.70} & .48           & .65 \\ \hline
        \textbf{\experiment{3.2}} & .68          & .68          & .46           & \textbf{.66} \\ \hline
        \textbf{\experiment{4.1}} & .57          & .66          & .58           & .58 \\ \hline
        \textbf{\experiment{4.2}} & .58          & .66          & \textbf{.59}  & .59 \\ \hline
    \end{tabular}}
    \caption{\gptTurbo, Spearman's Rank Correlation (\rhoSymbol). BG = \citet{greenberg2015improving}, B0 = \citet{tilk2016event}, B1 = \citet{hong2018learning}, B2 = \citet{Marton2021ThematicFB}, B3 = \citet{muthupari2022s} as baselines for our experiments. Subscripts a-c denote specific models there (Glove-shared-not-tuned, etc.). 
    \experiment{$m.n$} = see Table~\ref{tab:OverviewT}.
    All \rhoSymbol’s had p-values $< 10^{-13}$.}
    \label{tab:gpt4turbo}
\end{table}

\begin{table}[t!] 
     \resizebox{\columnwidth}{!}{
        \begin{tabular}{|p{3cm}|c|c|c|c|}
        \hline
        \diagbox{\textbf{Model}}{\textbf{Dataset}} & \textbf{\FerLoc} & \textbf{\FerIns}  & \textbf{Pado} & \textbf{McRae} \\ \hline
        \textbf{BG:}  GSD2015 & .29 & .42 & .53 & .36 \\ \hline
        \textbf{B0:} NN RF & .44 & .45 & .52 & .38 \\ \hline
        \textbf{B1:} ResRoFA-MT & .46 & .48 & .53 & .43 \\ \hline
        \textbf{B2:}  20\% subset v2 & - & - & .43 & .44 \\ \hline
        \textbf{B3a:}  Glove-shared-not-tuned & .30 & .34 & .49 & .31 \\ \hline
        \textbf{B3b:} Glove-shared-tuned & .29 & .29 & .53 & .33 \\ \hline
        \textbf{B3c:} Glove-shared-tuned with 20\%M & .35 & .43 &\textbf{.59}& .43 \\ 
        \hline
    
        \hline
        \textbf{\experiment{1.1}} & .51          & .56          & .40           & \textbf{.64}       \\ \hline
        \textbf{\experiment{1.2}} & .50          & .56          & .40           & .58 \\ \hline
        \textbf{\experiment{2.1}} & .32          & .33          & .33           & .40 \\ \hline
        \textbf{\experiment{2.2}} & .26          & .37          & .33           & .37 \\ \hline
        \textbf{\experiment{3.1}} & \textbf{.59} & \textbf{.59} & .45  & \textbf{.64 }\\ \hline
        \textbf{\experiment{3.2}} & .56          & \textbf{.59} & .42           & .63 \\ \hline
        \textbf{\experiment{4.1}} & .56          & .53          & .42           & .44 \\ \hline
        \textbf{\experiment{4.2}} & .53          & .53          & .42           & .46 \\ \hline
    \end{tabular}}
    \caption{\gptFourOne, Spearman's Rank Correlation (\rhoSymbol). 
    Legend is as in Table~\ref{tab:gpt4turbo}. All \rhoSymbol’s had p-values $< 10^{-5}$.}
    \label{tab:gptfourone}
\end{table}

\subsection{Open Models} \label{sec:Open Models}

The performance of the open models was lower than the closed models, and lower than several baselines across the board, except for \FerLoc\ using \stepByStepPrompting\ with \llama\ and McRae with \qwen\ (mostly in \experiment{4.x}). Details are below:

\paragraph{Axis 1: Reasoning Form}
Despite most results being lower than the baselines, \stepByStepPrompting\ (Exp3.x and 4.x) had a positive effect in \FerLoc\ and McRae. In Pado, the effect was detrimental with \llama\ (Table~\ref{tab:llama_3_2}), but positive with \qwen\ (Table~\ref{tab:qwen}).

\paragraph{Axis 2: Input Form}

Using Generated Sentences (Exp.2.x and 4.x) instead of Lemma Tuples 
had mostly no effect with \llama, except for gains in McRae.
With \qwen\ there were mostly gains except for lower scores in \experiment{4.x} in FerLoc.

\paragraph{Axis 3: Output Form}
Using predefined categories in the output, compared to numeric output (\experiment{x.2} vs. x.1), yielded lower scores with
\llama, except in Pado \experiment{3.x} and 4.x, but the correlation scores there are too low to trust these differences are meaningful. 
With \qwen\, results were worse in \FerIns\ and McRae, and mixed elsewhere.



\begin{table}[t] 
     \resizebox{\columnwidth}{!}{
        \begin{tabular}{|p{3cm}|c|c|c|c|}
        \hline
        \diagbox{\textbf{Model}}{\textbf{Dataset}} & \textbf{\FerLoc} & \textbf{\FerIns}  & \textbf{Pado} & \textbf{McRae} \\ \hline
        \textbf{BG:}  GSD2015 & .29 & .42 & .53 & .36 \\ \hline
        \textbf{B0:} NN RF & .44 & .45 & .52 & .38 \\ \hline
        \textbf{B1:} ResRoFA-MT & \textbf{.46 }& \textbf{.48} & .53 & .43 \\ \hline
        \textbf{B2:}  20\% subset v2 & - & - & .43 & \textbf{.44 }\\ \hline
        \textbf{B3a:}  Glove-shared-not-tuned & .30 & .34 & .49 & .31 \\ \hline
        \textbf{B3b:} Glove-shared-tuned & .29 & .29 & .53 & .33 \\ \hline
        \textbf{B3c:} Glove-shared-tuned with 20\%M & .35 & .43 &\textbf{.59}& .43 \\ 
        \hline
        \hline
        \textbf{\experiment{1.1}} & .32         & .33           & .27           & .16 \\ \hline
        \textbf{\experiment{1.2}} & .30         & .23           & .26           & .13 \\ \hline
        \textbf{\experiment{2.1}} & .32         & .33           & .31  & .25 \\ \hline
        \textbf{\experiment{2.2}} & .29         & .22           & .22           & .23 \\ \hline
        \textbf{\experiment{3.1}} & \textbf{.46}& .29  & .05*           & .23 \\ \hline
        \textbf{\experiment{3.2}} & .41         & .24           & .10*           & .20 \\ \hline
        \textbf{\experiment{4.1}} & \textbf{.46}& .28           & .06*           & .27 \\ \hline
        \textbf{\experiment{4.2}} & .41         & .24           & .12*           & .21 \\ \hline
    \end{tabular}}
    \caption{\llama, Spearman's Rank Correlation (\rhoSymbol).
    Legend is as in Table~\ref{tab:gpt4turbo}. All \rhoSymbol’s had p-values $< 10^{-3}$ except cells with * for which p-values were between [.01, .22].}
    \label{tab:llama_3_2}
\end{table}

\begin{table}[t!]
     \resizebox{\columnwidth}{!}{
        \begin{tabular}{|p{3cm}|c|c|c|c|}
        \hline
        \diagbox{\textbf{Model}}{\textbf{Dataset}} & \textbf{\FerLoc} & \textbf{\FerIns}  & \textbf{Pado} & \textbf{McRae} \\ \hline
        \textbf{BG:}  GSD2015 & .29 & .42 & .53 & .36 \\ \hline
        \textbf{B0:} NN RF & .44 & .45 & .52 & .38 \\ \hline
        \textbf{B1:} ResRoFA-MT & \textbf{.46} & \textbf{.48} & .53 & .43 \\ \hline
        \textbf{B2:}  20\% subset v2 & - & - & .43 & .44 \\ \hline
        \textbf{B3a:}  Glove-shared-not-tuned & .30 & .34 & .49 & .31 \\ \hline
        \textbf{B3b:} Glove-shared-tuned & .29 & .29 & .53 & .33 \\ \hline
        \textbf{B3c:} Glove-shared-tuned with 20\%M & .35 & .43 &\textbf{.59}& .43 \\ 
        \hline
        \hline
        \textbf{\experiment{1.1}} & .29  & .43 & .28  & .42 \\ \hline
        \textbf{\experiment{1.2}} & .38  & .37 & .19  & .34 \\ \hline
        \textbf{\experiment{2.1}} & .31  & .44 & .31  & .48 \\ \hline
        \textbf{\experiment{2.2}} & .37  & .38 & .22  & .39 \\ \hline
        \textbf{\experiment{3.1}} & .41  & .36 & .32  & .40 \\ \hline
        \textbf{\experiment{3.2}} & .41  & .32 & .36  & .35 \\ \hline
        \textbf{\experiment{4.1}} & .31  & .37 & .33  & \textbf{.51} \\ \hline
        \textbf{\experiment{4.2}} & .22  & .36 & .37  & .47 \\ \hline
    \end{tabular}}
    \caption{\qwen, Spearman's Rank Correlation (\rhoSymbol).
    Legend is as in Table~\ref{tab:gpt4turbo}. All \rhoSymbol’s had p-values $< 10^{-3}$.}
    \label{tab:qwen}
    \vspace{-2ex}
\end{table}

\section{Discussion} \label{Discussion}
The core questions we aim to address in this work are: Do \llms\ have the kind of linguistic knowledge required for TFE? If so, 
\textbf{(Q1)} does \stepByStepPrompting\   help utilize the model’s internal linguistics knowledge for this task? 
\textbf{(Q2)} Would providing more context by adding sentences to the input help? (as opposed to providing only lemma tuples and roles) 
\textbf{(Q3)} Do predefined thematic fit output categories better elicit the model's internal linguistic knowledge compared to numeric output?

Overall, we conclude that closed models acquired much of the linguistic knowledge required for TFE. Results with both GPT models on each of the test sets achieved or even surpassed all previous baselines, which were trained on labeled linguistic data (even if not directly annotated for TFE, such as SRL). This is impressive,\footnote{The available inter-annotator agreement or correlation information for our testsets indicates the upper bound for these tasks is between the mid-60's and mid-70's. \gptTurbo's scores got close, but no single setting was always the best.} assuming closed models were not trained on this task (sadly, we cannot know for sure because they are, well, closed).
Next, we discuss the findings by the three axes (as defined in \S\ref{sec:Experiments}) and then move to analyzing additional aspects.

\paragraph{(Q1) Axis 1: Reasoning Form}
For GPT models (closed models), \stepByStepPrompting\   fulfilled its intended goal of providing a “scratchpad” to help them calculate inferences given their internal knowledge (with the exception of \gptFourOne\ in Pado). In contrast, \stepByStepPrompting\ didn't help with open models (except for \FerLoc\ with \llama\ 
and for McRae with \qwen).

\begin{figure}
    \centering
    \resizebox{0.82\linewidth}{!}{
    \includegraphics[width=1\linewidth]{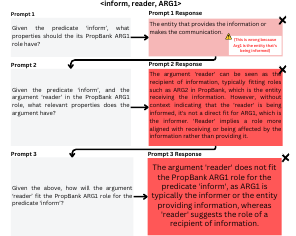}
    }
    \caption{Effect of Early Bad Reasoning. The example was taken from preliminary experimentation.}
    \label{fig:Effect of Early Bad Reasoning}
\end{figure}

Why was there a difference in the \stepByStepPrompting\ effect between the closed and open models?  \stepByStepPrompting\ is designed to help the models reason by first eliciting useful properties of arguments and roles, before the final TFE. However, our analysis on a sample of random tuples revealed that open models are weak at inferring the essential properties that are most expected for an argument, to serve well in a specific semantic role. Input from this step, containing invalid or irrelevant properties, often derailed the final reasoning output quality.
Moreover, if a reasoning step failed early on, it often derailed the rest of the chain (see Figure~\ref{fig:Effect of Early Bad Reasoning}) which may explain the score losses in some of the experiments.

 Why did \gptTurbo\ outperform \gptFourOne\ in spite of being older? We observed that \gptFourOne's ratings were often more `cautious', meaning it rarely output very high or very low scores. We verified this by calculating the variance of each model's output for each task and experimental setting, 
 finding a negative correlation between the variance and the correlation score (\rhoSymbol). The lower variance of \gptFourOne\ means it is more likely to do worse on inputs with very high or very low thematic fit.  

\paragraph{(Q2) Axis 2: Input Form}\label{'sec:5-axis}

In principle, providing more context enhances the models' ability to comprehend complex tasks. This was generally reflected in our experiments with open models, but not closed models (with few exceptions). 

Why do generated sentences only help open models?
Although the open models did not outperform the closed models, they often better leveraged the input with generated sentences, and improved the correlation score (\rhoSymbol) relative to the tuple input. But 
sentence input is likely to outperform tuple input only if these sentences are ``good'' (semantically coherent, with the given argument in the given role for the given predicate). Moreover, ``bad'' sentences are likely to derail the models' performance. Therefore, we analyzed a random sample of 20 tuples, 5 generated sentences for each, altogether 100 sentences per model. We found that the open \qwen\ did better in filtering out ``bad'' sentences (in spite of having fewer parameters): its semantic filtering let through no False Positives (FPs), thus successfully leveraging the sentence inputs. In contrast, the closed  \gptTurbo\ let through 6 FPs (Table~\ref{tab:error_comparison_tabularx}). We saw similar results comparing filtering by \qwen\ \textit{vs.} \gptTurbo, both on \gptTurbo's sentences.\footnote{Note that \qwen\ had higher False Negative rate, but this would only result in regressing to tuple input (\S\ref{sec:input} Axis 2).}
With it, closed models clearly generated more ``good'' sentences (80 vs. 60 True Positives + False Negatives in our sample). Therefore, we caution against jumping to conclusions which model gained more relevant knowledge. See also another potential factor in \S\ref{sec:perform-driver}.

\begin{table}
    \centering
    \resizebox{0.82\linewidth}{!}{
    \begin{small}    
    \begin{tabularx}{\columnwidth}{|p{2.2cm}|c|c|c|}
        \hline
        \textbf{\makecell{Acceptable\\Sentence?}} & 
        \textbf{\qwen} & 
        \textbf{\makecell{GPT4-\\Turbo}} & 
        \textbf{\makecell{\qwen\ filtering\\ \gptTurbo}} \\ 
        \hline
        True Positive & 18 & 52 & 26 \\
        \hline
        True Negative & 40 & 15 & 20 \\
        \hline
        False Positive & 0 & \textbf{6} & 1 \\
        \hline
        False Negative & 42 & 27 & 53 \\
        \hline
    \end{tabularx}    
    \end{small}}
    \caption{\footnotesize Semantic filtering of ``bad'' generated sentences, random sample size: 100. ``Bad'' = True Negatives + False Positives. Left 2 columns: model both generated and filtered.}
    \label{tab:error_comparison_tabularx}
    \vspace{-2ex}
\end{table}

\paragraph{(Q3) Axis 3: Output Form} 
There are arguments in favor of either categorical and numeric output.
\llms\ are primarily designed for text-based tasks, so we expect predefined output categories to yield better results than numeric output. However, \llms\ have also shown impressive performance in math (even if not perfect). Also, the ground-truth thematic fit scores were averaged across the participants, providing a graded scale, and therefore perhaps a numeric output would better correlate with that. 
Indeed, output form had almost no effect for closed models, but categorical output gave mostly worse results with open models. 
We hypothesize this may be due to the differences in post-training between the closed and open models, but we currently have no solid explanation for this difference, and we leave this for future work.

\subsection{Performance drivers\label{sec:perform-driver}\protect\footnote{We thank the anonymous reviewers for motivating us to add this subsection. OpenAI retired \textit{gpt-4-0125-preview} after we used it for the main results (Tables \ref{tab:gpt4turbo}-\ref{tab:error_comparison_tabularx}), so for all GPT runs here (Tables \ref{tab:ablation1}-\ref{tab:model_improve_comparison}), including the baselines, we use \textit{gpt-4-turbo-2024-04-09}, the closest available successor.}}

What is the main performance driver? \textbf{(a)} Model knowledge (model size and specific weights), \textbf{(b)} user-provided knowledge (system prompt context), or \textbf{(c)}  generated sentences (input context) quality? 

Model size (a) is a prime suspect, and is a confound with open vs. closed models (the latter being x200 larger). But 
\qwen\  outperforming \gptTurbo\ 
on filtering ``bad'' sentences is a counter-example.
While we cannot fully interpret and compare model weights, we can explore (b) and (c).

\paragraph{System prompt knowledge} \label{ab-1}
We tested  removing semantic role definitions from the system prompt for \gptTurbo\ (Table~\ref{tab:ablation1}). As expected, \gptTurbo\ did worse on \FerIns\  when \role{Instrument} definitions were absent, but did surprisingly better on McRae  when \role{Agent, Patient} and related information was removed (especially in \experiment{3.1}). It is doubly surprising, as adding role definitions during our preliminary studies (with the older GPT model) improved performance by 8-9 points in \experiment{3.x}.
We suspect fundamental differences between the 2 GPTs are the cause, and wish these models were open.
McRae output inspection reveals mixed effects: sometimes without role definitions GPT misinterpreted what properties the PropBank semantic role should have; other times including role definitions made GPT assign low score for an object lacking an optional property (e.g., low fit for `rabbit' in \role{Patient/Arg1} role due to not being stationary).

\begin{table}
\centering
\footnotesize
\setlength{\tabcolsep}{3pt}
\begin{tabular}{lcc|cc}
\hline
& \multicolumn{2}{c|}{\textbf{\FerIns}} 
& \multicolumn{2}{c}{\textbf{McRae}} \\
\cline{2-5}
\textbf{Exp.} 
& \textbf{$r$ (WR $\rightarrow$ WoR)}
& \textbf{$\Delta r$}
& \textbf{$r$ (WR $\rightarrow$ WoR)}
& \textbf{$\Delta r$} \\
\hline
\textbf{3.1} 
& .721 $\rightarrow$ .718 & -.003
& .621 $\rightarrow$ .649 & \textbf{.028} \\

\textbf{3.2} 
& .718 $\rightarrow$ .709 & -.009
& .630 $\rightarrow$ .635 & .005 \\
\hline
\end{tabular}
\caption{ Ablation test: correlation ($r$) with and without role definition in system prompt. Exp. = experiment, WR = with roles, WoR = without roles, $\Delta$ = WoR-WR.}
\label{tab:ablation1}
\end{table}

\paragraph{Generated Sentence Quality}
Was the sentences' generation and/or filtering quality a major factor?
To assess this, 
we reused the $100$-sentence \FerIns\ sample  (\S\ref{'sec:5-axis}, Axis~2) and manually fixed false positives (by providing a compliant sentence) and false negatives (by using the wrongly filtered out sentence), for both \gptTurbo\ and \qwen\  (Table~\ref{tab:model_improve_comparison}).
%

%
We did not see evidence for the sentences being a large factor: the effect of fixing the sample sentences was tiny (mostly $\pm$.005),
even though the sample size was 8\% of the dataset. 

\begin{table}
\centering
\small
\setlength{\tabcolsep}{3pt}
\begin{tabular}{lcc|cc}
\hline
& \multicolumn{2}{c|}{\textbf{\gptTurbo}} 
& \multicolumn{2}{c}{\textbf{\qwen}} \\
\cline{2-5}
\textbf{Exp.} 
& \textbf{$r$ (Base $\rightarrow$ Good)}
& \textbf{$\Delta r$}
& \textbf{$r$ (Base $\rightarrow$ Good)}
& \textbf{$\Delta r$} \\
\hline
\textbf{2.1} 
& .422 $\rightarrow$ .427 & { }.005
& .439 $\rightarrow$ .444 & { }.005 \\

\textbf{2.2} 
& .509 $\rightarrow$ .505 & -.004
& .379 $\rightarrow$ .370 & -.009 \\
\hline
\end{tabular}
\caption{Using ``good'' sentences: manually fixing false positives and false negatives in a random sample.  Correlation ($r$) for \gptTurbo\ and \qwen\ on \FerIns, $\Delta$ = Good-Base.}
\label{tab:model_improve_comparison}
\end{table}

\begin{figure}
    \centering
    \includegraphics[width=1.0\linewidth]{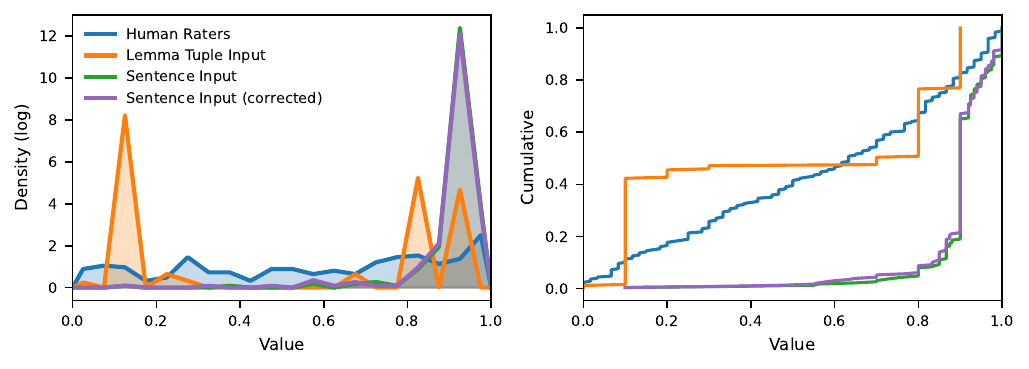}
    \caption{\FerIns\ TFE distributions of human raters (almost uniform), tuple input (bipolar), sentence input and corrected sentences (both right-heavy).}
    \label{fig:distribs}
\end{figure}

These small (and sometimes negative) effects made us look further: 
Recall that instead of every filtered sentence, we back off to using the score from the counterpart tuple condition. But if tuple vs. sentence input result in very different score distributions, this back-off can lead to unexpected results!
Indeed, Figure~\ref{fig:distribs} shows that human scores are almost uniform (somewhat heavier near the high end), lemma tuple input scores are bipolar around .1 and .8-.9, sentence input scores are almost all above .8, where manually correcting the random sample sentences had a small smoothing effect between .9-1.0. Recall also that higher TFE scores do not necessarily mean better correlation with the ground truth. These findings are interesting for themselves, and should guide future work.

\section{Conclusion and Future Work}

We set out to discover if, or to what extent, autoregressive \llms\  possess linguistic and other knowledge required to estimate the thematic fit of event participants. 
The tested \llms\ set
new state-of-the-art  on each test set (and tied on Pado). 
Still,
we conclude that TFE is not yet ``solved''. Closed models\footnote{We observed that \gptFourOne\ did worse than \gptTurbo, in spite of being newer, likely because 4.1 is too `cautious' (\S\ref{Discussion}).} yielded higher results than open models, indicating that closed models, especially \gptTurbo, acquired much of the linguistic knowledge required for TFE. 
This could be due to model size, training data and/or training regime.
But open models were better at filtering ``bad'' generated sentences, 
and our back-off  may have also greatly affected results,
so we caution against rushing to interpret this as showing  open models gained less relevant knowledge.
No single axis setting consistently yielded the highest scores, so it is still unclear if models need additional knowledge or a different elicitation.


We tested three axes for eliciting this knowledge: \textbf{(1) Reasoning Form:} adding \stepByStepPrompting, \textbf{(2) Input Form:} adding generated sentences as context, and \textbf{(3) Output Form:} predefined categories vs. numeric scores. 

\textbf{(1)} Surprisingly, 
\stepByStepPrompting\ only helped closed models; Our analysis showed that invalid or irrelevant reasoning 
not only was unhelpful but actually hurt performance. This happened more in open models.

\textbf{(2)} Generated sentences hurt closed models; We saw \llms\ struggle with generating sentences where the given argument is a better fit to a semantic role different than the given one. To tackle this, we used novel semantic filtering, which, to our surprise, worked better for open models, because they were better at filtering out False Positives. 

\textbf{(3)} Predefined categorical output (vs. numeric output) had zero-to-small effect with closed models and more negative effect with open models. 

In future work, following our analysis, we hope to improve \stepByStepPrompting,     generate better sentences, better filter ``bad'' sentences, and better instruct the models to output more balanced estimation. If successful, \llms\ may also be used to study linguistic issues in thematic fit, in line with \citet{futrell2025how}, BabyLM efforts\footnote{\url{https://babylm.github.io}}, and others who believe that \llms\ can be useful tools to study linguistic phenomena.

\section*{Limitations} 
Our experiments were evaluated on small datasets (a known limitation of this computational psycholinguistics subfield, as opposed to typical machine learning), so findings and conclusions should be assessed with caution.
Our study considered only few \llms, so one should be cautious about drawing conclusions for all \llms.

\llms\ can be very sensitive to prompt phrasing, so it is possible that different prompts may yield very different results. Also, prompt engineering is tricky; thus, tweaking the prompts may lead to different results. The richness of prompt variation (prompt engineering) depends on human resources (different people use language differently) and funding/time resources, all of which were limited. 
Future work may achieve better results with more resources, if available. 
Due to the scarcity of evaluation data, we could not use prompt optimization methods such as DSPy \cite{Khattab2023DSPyCD} or fine-tune the models.

Last, all findings and conclusions here rely on data and annotations in English, as it is by far the most resource-rich. But thematic fit (and semantics in general) are thought of as universal, and of course relevant to all languages. Still, we would advise to be cautious of drawing universal conclusions before validating our findings on an additional, diverse set of languages.

\section*{Acknowledgments}
The first author would also like to express gratitude to her sponsors, the Kuwait Chamber of Commerce and Industry and Abdullah Al Salem University, for granting her scholarships that made this research possible at Columbia University and Imperial College London.

\bibliography{custom.bib}

\appendix
\section*{Appendices}

\section{Prompt Design} \label{sec:ApdxPromptDesign}
Each experiment specified in \S\ref{sec:ExperimentDesign} has a prompt design with specific input, output, and reasoning:

\paragraph{\simplePrompting\  with Lemma Tuples \small{(\experiment{1.1}--\experiment{1.2})}}
A single prompt template (`simple thematic fit') that directly asks for the thematic fit score of a  $\langle$predicate, argument, role$\rangle$ tuple (Fig.~\ref{fig:overviewExp}). 

\begin{itemize}[itemsep=2pt, topsep=2pt, parsep=0pt, partopsep=0pt, leftmargin=*]
\item \prompt{Given the predicate ‘eat’, how well does the argument ‘pizza’ fit the PropBank \argOne\ role?} 
\end{itemize}

We extend the prompt with two output settings:

\begin{itemize}[itemsep=2pt, topsep=2pt, parsep=0pt, partopsep=0pt, leftmargin=*]
    \item x.1: Numerical score where the model provides a numeric value between 0 and 1 to represent the thematic fit score: \newline
    \prompt{Reply only with a valid JSON dictionary \{"Score": numeric\_value\} where numeric\_value is a float number from 0 to 1. Avoid adding any text outside this JSON dictionary.}
    \item x.2: Categorical score where the model selects one of the predefined labels that indicate the thematic fit score: \newline
    \prompt{Reply only with a valid JSON dictionary \{"Score": string\_value\} where string\_value that is one of "Perfect", "High", "Medium", "Low" or "Unlikely/Impossible". Avoid adding any text outside this JSON dictionary.}
\end{itemize}

\paragraph{Sentence Generation and Semantic Filtering}\label{sec:gensentenceApx}

Before executing experiments 2 and 4, the model is asked to generate $k$ diverse semantically coherent sentences containing an argument in a specific semantic role, given a predicate. Here we use $k=5$.
\begin{itemize}[itemsep=2pt, topsep=2pt, parsep=0pt, partopsep=0pt, leftmargin=*]
    \item \prompt{Give me five sentences with the predicate ‘eat’ and argument ‘pizza’ in the PropBank \argOne\ role. Make the sentences as diverse as possible, while using the predicate as the main verb and keeping the given argument in the specified role. Make sure that each sentence is semantically coherent and the argument is in the given role.}
\end{itemize}
\hfill
\llms\ may often generate a semantically incoherent or incompatible sentence with respect to the given role for the argument. Thus, we design a semantic filtering process of 4 waterfall Yes/No Steps. If the sentence passed the whole waterfall at least once out of three trials, it is considered semantically acceptable. Here are the prompts we used:

\begin{itemize}[itemsep=2pt, topsep=2pt, parsep=0pt, partopsep=0pt, leftmargin=*]
    \item \textbf{(2a)} \prompt{Here’s a sentence: ‘I ate a pizza with my friends’. Is the given sentence semantically coherent?} 
    \item \textbf{(2b)} \prompt{For the sentence: ‘I ate a pizza with my friends’. Does the sentence contain the predicate ‘eat’?} 
    \item \textbf{(2c)} \prompt{For the sentence: ‘I ate a pizza with my friends’. Does the sentence contain ‘pizza’ as an argument of the predicate ‘eat’? } 
    \item \textbf{(2d)} \prompt{For the same sentence: ‘I ate a pizza with my friends’. Given the definitions, instructions, and answers above, is ‘pizza’ in the PropBank \argOne\ role for the predicate ‘eat’?} 
\end{itemize}
\hfill

\paragraph{\simplePrompting\ with Generated Sentences \small{(\experiment{2.1}--\experiment{2.2})}}
We augment the prompt in \experiment{1.x} with sentences generated by the model from the same lemma tuples:
\begin{itemize}[itemsep=2pt, topsep=2pt, parsep=0pt, partopsep=0pt, leftmargin=*]
    \item \prompt{Given the following sentence, ‘I ate a pizza with my friends’, how well does the argument ‘pizza’ of the predicate ‘eat’ fit the PropBank \argOne\ role?} 
\end{itemize}
\hfill

\paragraph{\stepByStepPrompting\   with Lemma Tuples \small{(\experiment{3.1}--\experiment{3.2})}}

Experiments 3.x are the enhanced version of \experiment{1.x}, where we break down the thematic fit task into a series of 4 prompts (`\stepByStepPrompting'). See Fig.~\ref{fig:overviewExp}.

\begin{itemize}[itemsep=2pt, topsep=2pt, parsep=0pt, partopsep=0pt, leftmargin=*]
    \item \prompt{What salient essential properties or core characteristics best describe the essence of ‘pizza’?}\footnote{In an early version of this experiment, only the semantic role name was mentioned. After analyzing it, we suspected that the \llms\ may have not made the connection between roles such as ‘\argOne’ and their PropBank sense in this context. Therefore we added the word ‘PropBank’ before the semantic role for  Pado and McRae, as their semantic roles were PropBank-based, unlike the Ferretti datasets. This indeed improved the results on these sets by 2-3\% (absolute). 
    } 
    
    \item \prompt{Given the predicate ‘eat’, and the argument of it in the PropBank \argOne\ role, what salient essential properties or core characteristics are most expected for this argument, to serve well in the PropBank \argOne\ role?}

    \item \prompt{Given the predicate ‘eat’, what PropBank roles are most suitable or likely for its argument ‘pizza’?}

    \item \prompt{Given the predicate ‘eat’ with the argument ‘pizza’ in the PropBank \argOne\ role, what essential properties does the argument have, properties which fit well its PropBank \argOne\ role? And what essential properties this argument should have had (but are missing) in order for it to serve well in its PropBank \argOne\ role?}
\end{itemize}
\hfill

\paragraph{\stepByStepPrompting\   with Generated Sentences \small{(\experiment{4.1}--\experiment{4.2})}}
This experiment uses \stepByStepPrompting\ as in \experiment{3.x} but with the addition of the generated sentence as in \experiment{2.x}:

\begin{itemize}[itemsep=2pt, topsep=2pt, parsep=0pt, partopsep=0pt, leftmargin=*]
    \item \prompt{Here’s a sentence:  ‘I ate a pizza with my friends.’, What salient essential properties or core characteristics best describe the essence of ‘pizza’?}
\end{itemize}

\section{Earlier round of Experiment}
A previous round of these experiments has been done on \gptTurbo\ and CodeLlama2 (codeLlama2-70B-instruct) \citep{roziere2023code} with a previous version of the prompts we use. The results of the earlier round 
were mixed: sometimes higher (up to .77 on \FerLoc) and other times lower than our new prompts. We take it as a positive sign that we didn't overfit the new prompts.


\end{document}